\documentclass[conference]{IEEEtran}
\IEEEoverridecommandlockouts
\usepackage{cite}
\usepackage{amsmath,amssymb,amsfonts}
\usepackage{algorithmic}
\usepackage{graphicx}
\usepackage{textcomp}
\usepackage{xcolor}
\usepackage{fancyhdr} 
\usepackage{lipsum} 
\def\BibTeX{{\rm B\kern-.05em{\sc i\kern-.025em b}\kern-.08em
    T\kern-.1667em\lower.7ex\hbox{E}\kern-.125emX}}

\fancypagestyle{firstpageheader}{
    \fancyhf{} 
    
}  
\begin{document}

\title{Real-time Bangla Sign Language Translator

}

\author{\IEEEauthorblockN{Rotan Hawlader Pranto}
\IEEEauthorblockA{\textit{Department of Electrical and Computer Engineering} \\
\textit{North South University}\\
Dhaka, Bangladesh \\
rotan.pranto@northsouth.edu}
\and
\IEEEauthorblockN{ Shahnewaz Siddique}
\IEEEauthorblockA{\textit{Department of Electrical and Computer Engineering} \\
\textit{North South University}\\
Dhaka, Bangladesh \\
shahnewaz.siddique@northsouth.edu}
\and
}

\maketitle
\thispagestyle{firstpageheader}
\begin{abstract}
The human body communicates through various meaningful gestures, with sign language using hands being a prominent example. Bangla Sign Language Translation (BSLT) aims to bridge communication gaps for the deaf and mute community. Our approach involves using Mediapipe Holistic to gather key points, LSTM architecture for data training, and Computer Vision for realtime sign language detection with an accuracy of 94\%.

\end{abstract}
\renewcommand\IEEEkeywordsname{Keywords}

\section{Introduction}
Communication is essential for expressing feelings, yet it poses significant challenges for the deaf community. Globally, there are approximately 466 million deaf individuals, including 36 million children. In Bangladesh alone, around 13.7 million people are deaf. Disabilities affect approximately 16\% of the world's population, underscoring the importance of inclusive communication solutions.

Sign language is the most efficient way to make communication between deaf and dumb people. However, this is only possible if both of them have an acute knowledge of sign language. Each word and alphabet has specific signs. So, prior knowledge is needed to facilitate convenient communication. Fig 1 illustrates different signs which represent the alphabets of the Bangla Sign language.

\begin{figure}[htbp]
    \centering
    \includegraphics[width=0.4\textwidth]{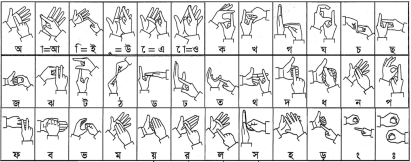}
    \caption{ Bangla alphabet sign [7]}

\end{figure}

Deep learning encompasses diverse methodologies, among which RNNs  stand out for their effectiveness in processing sequential data. Long Short-Term Memory networks, a specialized type of RNN, are particularly renowned for their ability to manage and predict data sequences over time. In the context of BSLT , LSTM networks are used for training due to their proficiency in handling temporal patterns. The system integrates computer vision techniques for real-time sign language detection, utilizing the Mediapipe library to collect and track key points and landmarks on the user's hands and face. The PIL renders fonts, ensuring clear and accurate visual representation of translated text. By combining these advanced frameworks and libraries, BSLT efficiently translates Bangla sign language into written text, facilitating user communication and enhancing accessibility.

\section{ RELATED WORKS}
Previously good number of work has been done to recognize and translate sign language using various methods and technologies using different types of sensors. Among them, some relevant works are explained below.

\subsection{Neural Sign Language Translation [1]}

This paper offers a concise overview of sign language translation methodologies. The researchers utilized the PHOENIX-weather 2014T dataset to train their model. They achieved an high level translation performance of 19.26 BLEU-4. Their frame level and gloss level tokenization network achieved respective scores of 9.85 and 18.13.
\begin{figure}[htbp]
    \centering
    \includegraphics[width=0.4\textwidth]{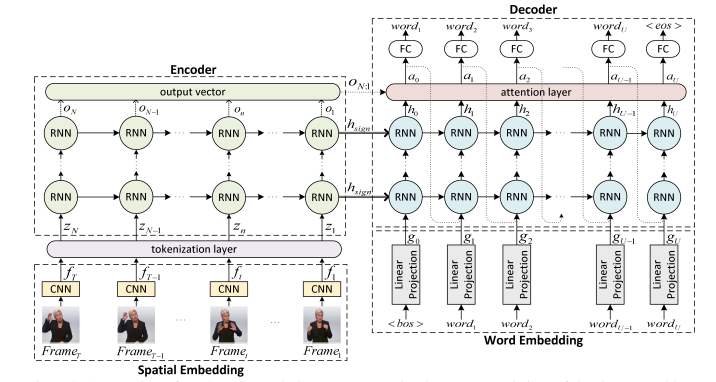}
    \caption{ An overview of SLT approach .[1]}

\end{figure}

\subsection{Real-time Sign Language Recognition using Computer Vision [2]}
Their paper addresses the societal gap between differently-abled individuals, such as those who are deaf and mute, and others. Image processing techniques are employed to preprocess images and extract hands from backgrounds effectively. The researchers utilized CNN to evaluate their custom dataset and real-time hand gestures, achieving an accuracy of 83\%.

\subsection{ Research of a Sign Language Translation System Based on Deep Learning [3] }

Their paper explores hand localization and sign language detection using neural networks. The approach integrates faster R-CNN for sign detection, 3D CNN for feature extraction, and LSTM networks for sequence encoding and decoding. The study achieved a notable 99\% accuracy in recognizing common vocabulary datasets for sign language.

\begin{figure}[htbp]
    \centering
    \includegraphics[width=0.4\textwidth]{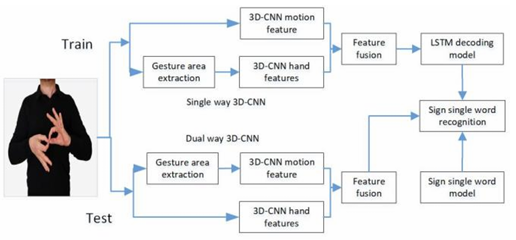}
    \caption{ 3D CNN and LSTM Encoding and Decoding Structure[3]}
\end{figure}
\subsection{ Continuous Sign Language Recognition with Correlation Network [4] }
This paper (CorrNet) improves CSLR by capturing body trajectories across frames using a correlation module. It achieves significant accuracy on datasets including PHOENIX14, with a Word Error Rate of 18.8\% on the training set and 19.4\% on the test set. CorrNet's effectiveness is demonstrated through comprehensive comparisons and visualizations, highlighting its ability to strengthen human body movements across adjacent frames.

\begin{figure}[htbp]
    \centering
    \includegraphics[width=0.4\textwidth]{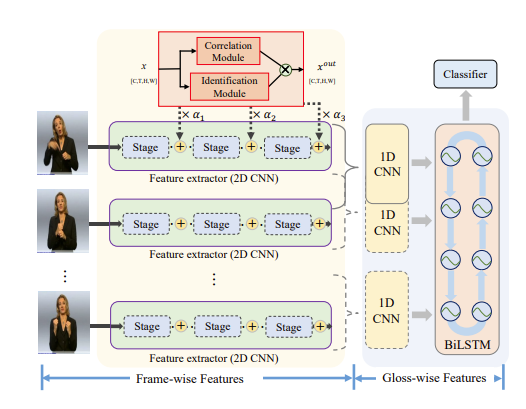}
    \caption{ An overview of CorrNet [4]}
\end{figure}

\subsection{ Sign Language Translator and Gesture Recognition [5] }
The paper explores a sign language translator and gesture recognition system. The system leverages deep learning and computer vision techniques to accurately translate sign language and gestures. They have built a smart glove to detect hand movements. The proposed model demonstrates an accuracy score of 96\%, showcasing its effectiveness in facilitating communication for individuals who use sign language.
\begin{figure}[htbp]
    \centering
    \includegraphics[width=0.4\textwidth]{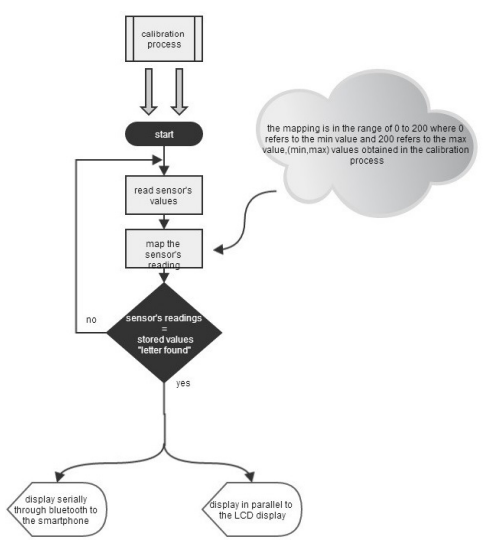}
    \caption{ The testing procedure flowchart [5]}
\end{figure}

\section{ PROPOSED SOLUTION}
\subsection{ Recurrent neural network}
Recurrent neural networks (RNNs), particularly those featuring LSTM architecture, are renowned for their ability to handle and generate sequential data. This makes them highly useful in applications such as sign language translation, where the sequential nature of gestures and signs are crucial. BSLT leverages LSTM networks for both encoding and decoding sequences of sign language gestures, allowing for the formulation of coherent and contextually accurate sentences. The system's integration of real-time sign detection ensures that the most recent and accurate signs are used to generate translations. By maintaining context over time and effectively processing sequential data, (BSLT) significantly enhances meaningful communication for individuals who use sign language, bridging gaps and facilitating better interactions. The advanced capabilities of LSTM networks in managing temporal dependencies is instrumental in achieving high accuracy and reliability in Bangla sign language translation.

\begin{figure}[htbp]
    \centering
    \includegraphics[width=0.4\textwidth]{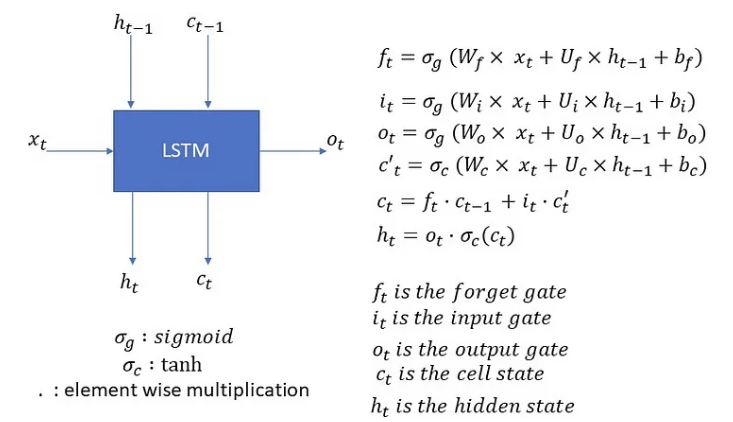}
    \caption{  Long short-term memory [11]}
    
\end{figure}

\subsection{Methodology}
We utilized Mediapipe for extracting key points and landmarks detection from images. The collected data was processed by flattening the key points into NumPy arrays and appropriately labeled. Our model employed an LSTM architecture to achieve high accuracy in generating coherent sentences, Where we used three LSTM layer, Two dense layer and relu activation function. Additionally, we utilized the PIL library to render Bangla fonts, ensuring linguistic accuracy. Real-time execution was facilitated through computer vision.

\begin{figure}[htbp]
    \centering
    \includegraphics[width=0.3\textwidth]{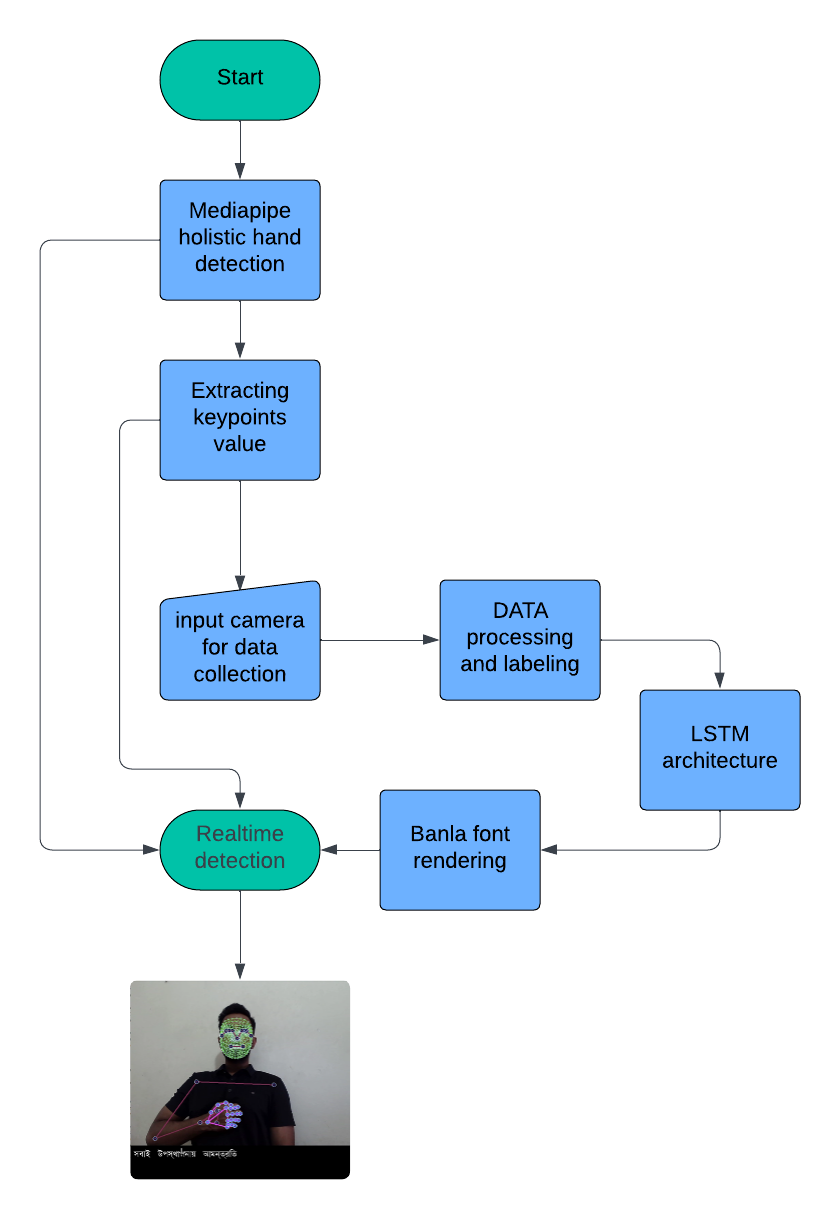}
    \caption{Methodology}
    
\end{figure}

\subsection{Data Collection}
We systematically collected sequential data for each word, capturing 30 frames per word using MediaPipe Holistic for comprehensive landmark detection, including hands, face, and pose. Key points were extracted and flattened into NumPy arrays, ensuring a structured representation of sign language gestures for effective processing and analysis The Bangla sign language translator.
\begin{figure}[htbp]
    \centering
    \includegraphics[width=0.39\textwidth]{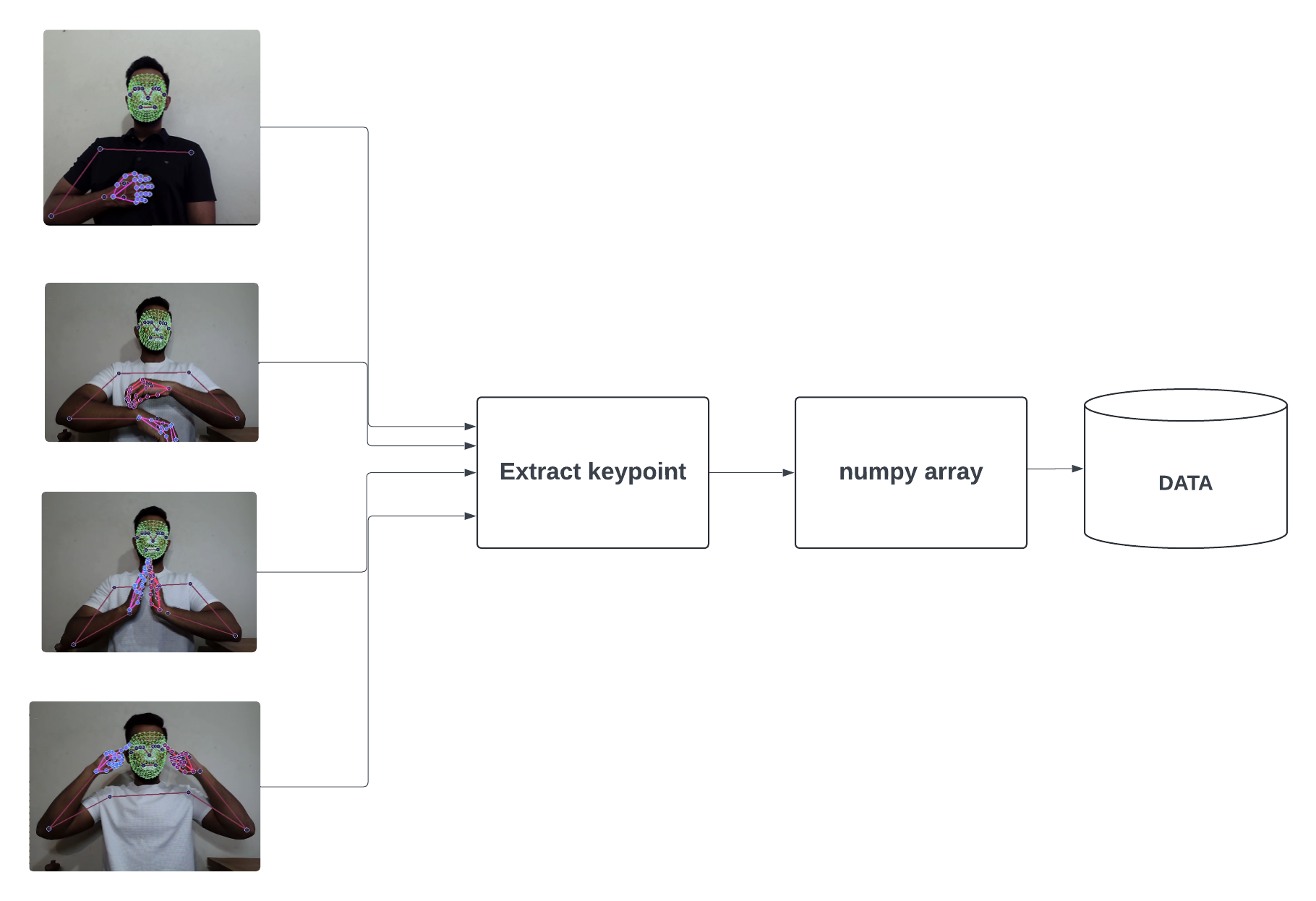}
    \caption{Data collection}
    
\end{figure}
\subsection{Results}
Collecting data for Bangla Sign Language presented significant challenges, particularly due to the need for accurate and diverse representation of signs. Despite these difficulties, we successfully gathered and meticulously labeled the data, capturing 30 frames per word to ensure comprehensive coverage. We employed a Long Short-Term Memory (LSTM) architecture for training, achieving a commendable accuracy score of 94\% and an F1 score of 93\%. The model’s performance was further validated through various tests by checking its effectiveness in real-life tasks. Our results include a detail presentation of the model predictions, as well as a screenshot showcasing real-time detection capabilities. These demonstrations highlight the practical utility of the model, offering a robust solution for real-time sign language recognition. The successful implementation of this model illustrates its potential for enhancing communication and accessibility for Bangla sign language users.
\begin{table}[htbp]
\caption{In this table, we have shown the results}
\centering
\begin{tabular}{ |c|c|c| } 
 \hline
 Architecture & Accuracy & F1 Score \\ 
 \hline
 LSTM & 94\% & 93\% \\ 
 \hline
\end{tabular}
\end{table}

\begin{figure}[htbp]
    \centering
    \includegraphics[width=0.45\textwidth]{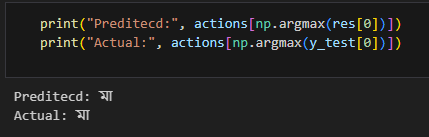}
    \caption{Predicted result}
    
\end{figure}
\begin{figure}[htbp]
    \centering
    \includegraphics[width=0.45\textwidth]{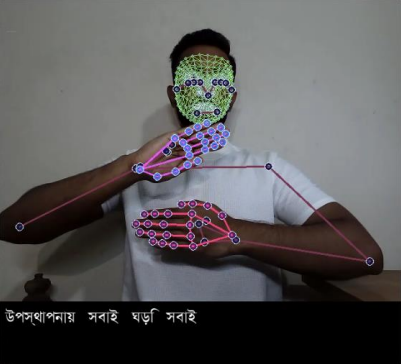}
    \caption{ Realtime Detection}
    
\end{figure}
\subsection{Experimental setup}
Our project has utilized several libraries and tools, which includes OS, LSTM, MediaPipe, Matplotlib, scikit-learn, OpenCV, TensorFlow, TensorBoard, NumPy and Metrics. Additionally, a camera is used to test and train the model. The hardware configuration includes an AMD Ryzen 7 5700X CPU, a GeForce RTX 3060 OC 12GB GPU, 16 GB of 3200 MHz RAM, and an MSI B550 GEN3 motherboard.
\section{ Analysis}
The most challenging part we faced while doing the project was the deficiency of an extensive Bangla dataset for each alphabet and words, which initially created some difficulties on data collection. Moreover, distinguishing between the signs for “deer” and “educated” showed some detection challenges. However, we resolved the problem before labeling and starting the training process. Additionally, we faced difficulties with rendering Bangla fonts, as they were not displaying correctly, which led to errors in visual representation. Initially, our model exhibited low accuracy during the early training phases. However, after 500 epochs of training using LSTM, we overcame fluctuations in accuracy observed between epochs 310 and 330, ultimately achieving an impressive accuracy rate of 94\%. Concurrently, our loss function showed a steady decline, reaching minimal values by the 500th epoch. This stability in both accuracy and loss signifies the effectiveness of the model’s learning process. For a visual representation of our progress, please refer to the training accuracy and loss progression graph provided below. The successful resolution of these challenges highlights the robustness of our approach and its effectiveness on real life applications. Additionally, ongoing improvements to the dataset and font rendering are expected to further enhance the model’s performance and usability, paving the way for broader adoption and integration in assistive technologies.
\begin{figure}[htbp]
    \centering
    \includegraphics[width=0.45\textwidth]{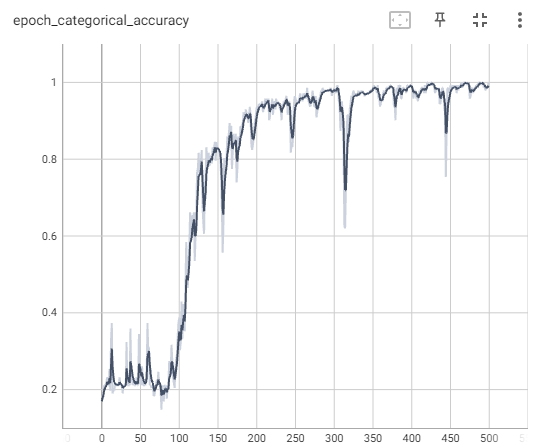}
    \caption{ Accuracy graph}
    
\end{figure}

\begin{figure}[htbp]
    \centering
    \includegraphics[width=0.45\textwidth]{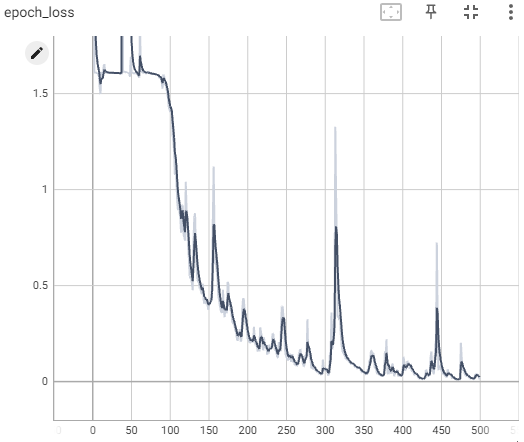}
    \caption{Training loss graph}
    
\end{figure}

\section{ Impact of BSLT}
BSLT, an AI-powered innovation, is set to significantly impact third-world countries like Bangladesh, especially within the hearing-impaired community. By dismantling traditional communication barriers, BSLT aims to more fully integrate the deaf community into society. This advancement harnesses AI’s transformative potential to enhance access to education, employment, and social inclusion. Moreover, BSLT represents a crucial step toward fostering equality and improving the quality of life of a individual with hearing difficulty. By paving the way for greater inclusivity, it contributes to a more equitable future and empowers individuals to participate more actively in societal development.

\section{ CONCLUSION}
Real-time Bangla Sign Language Translation (BSLT) aims to bridge communication gaps for the deaf and mute community. Integration of MediaPipe for sign detection and RNN for sequential result empowered the development of this system. Additionally, including computer vision (CV2) took the model to the next level. Real-time translations created a convenient and innate communication for the users. Advancement of Deep learning technologies, specially in sign language system is breaking down the communication barriers and enabling more inclusive interactions. More potential outcome can be generated by future innovations. 
\begin{table}[htbp]
\caption{ACCURACY COMPARISON WITH RELATED WORKS}
\centering
\begin{tabular}{ |c|c|c| } 
 \hline
 Reference & Architecture & Accuracy \\ 
 \hline
 Our study & LSTM & 94\% \\ 
  \hline
  Reference-[3]  &  LSTM & 91.6\% \\ 
 \hline
  Reference-[3] & 3DResNet18+SVM-fusion & 98.3\% \\ 
  \hline
  Reference-[6] & Faster R-CNN  & 98.20\% \\ 
  \hline

\end{tabular}
\end{table}

\section{  FUTURE SCOPE}
Adding the ability to translate sign language to voice is a major breakthrough that will make communication with people who are deaf or hard of hearing more convenient and realistic. This upgrade will increase the confidence of the deaf and blends directly into everyday conversations. Integrating Natural Language Processing(NLP) and computer vision to unleash it's full potential in the sector with features such as sentence auto-completion and predictive text. Not only will this set unprecedented standards in accessibility and integration, it will also pave the path for future breakthroughs in sign language technology. With growing simplicity and precision in a range of use cases, we will take this step in the future.

Subsequently, integration of ML algorithms with deep learning architectures can elaborate significant improvement on the sign language translation models. Collaborative efforts with deaf and blind community will play a crucial role for data collection and training process. It will help to interpret the gestures more accurately. The potential for real time translation in educational institutes, workplaces, and social interactions are immense, promising to bridge communication gaps and foster a more inclusive society. Moreover, Fine tuning the model by users feedback and interactive development, a robust and user friendly sign language translation system can be created.

\section*{Acknowledgment}

I am extremely grateful to Dr. Shahnewaz Siddique for his unwavering supervision and invaluable guidance throughout the project. His expertise and commitment have played a vital role on the success of this project. I also extend my heartfelt gratitude to everyone who has supported and contributed throughout this journey.

\vspace{12pt}
\end{document}